\documentclass[letterpaper]{article} % DO NOT CHANGE THIS
\usepackage{aaai2026}  % DO NOT CHANGE THIS
\usepackage{times}  % DO NOT CHANGE THIS
\usepackage{helvet}  % DO NOT CHANGE THIS
\usepackage{courier}  % DO NOT CHANGE THIS
\usepackage[hyphens]{url}  % DO NOT CHANGE THIS
\usepackage{graphicx} % DO NOT CHANGE THIS
\usepackage{natbib}  % DO NOT CHANGE THIS AND DO NOT ADD ANY OPTIONS TO IT
\usepackage{caption} % DO NOT CHANGE THIS AND DO NOT ADD ANY OPTIONS TO IT
\usepackage{algorithm}
\usepackage{algorithmic}

\usepackage{colortbl}
\usepackage{xcolor}
\usepackage{graphicx}   
\usepackage{capt-of}

\usepackage{xcolor}
\usepackage{overpic}

\usepackage{amsmath} % Added for mathematical environments and symbols
\usepackage{amssymb} % Added for additional symbols, e.g., \mathbb
\usepackage{amsfonts} % May be needed for certain math fonts

\usepackage{siunitx}
\usepackage{booktabs}      
\usepackage{makecell}      

\usepackage{newfloat}
\usepackage{listings}
\DeclareCaptionStyle{ruled}{labelfont=normalfont,labelsep=colon,strut=off} % DO NOT CHANGE THIS
\floatstyle{ruled}
\newfloat{listing}{tb}{lst}{}
\floatname{listing}{Listing}

\title{Bridging Day and Night: Target-Class Hallucination Suppression in Unpaired Image Translation}
\author {
	Shuwei Li\textsuperscript{\rm 1},
	Lei Tan\textsuperscript{\rm 1},
	Robby T. Tan\textsuperscript{\rm 1, \rm 2}
}
\affiliations {
	\textsuperscript{\rm 1} National University of Singapore\\
	\textsuperscript{\rm 2} ASUS Intelligent Cloud Services (AICS)\\
	shuwei@u.nus.edu, lei.tan@nus.edu.sg, robby.tan@nus.edu.sg
}

\usepackage{bibentry}
\begin{document}

\maketitle
	\begin{figure*}
	\centering
	\includegraphics[width=\textwidth]{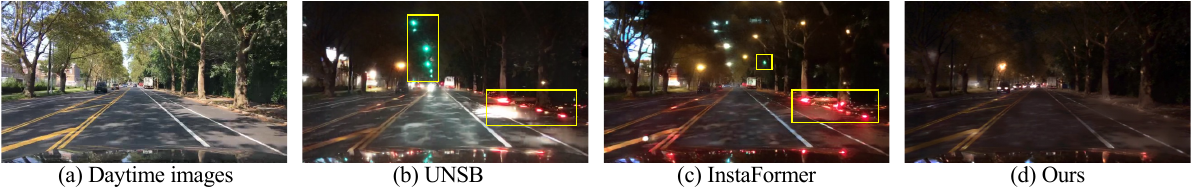}
	\caption{Existing unpaired image-to-image (I2I) translation methods often introduce hallucinations that mimic or falsely suggest target classes, such as fake taillights for cars or spurious green and red signals for traffic lights (see yellow boxes). 
	}
	\vspace{-4mm}
	\label{fig:teaser}
\end{figure*}

\begin{abstract}
Day-to-night unpaired image translation is important to downstream tasks but remains challenging due to large appearance shifts and the lack of direct pixel-level supervision.
Existing methods often introduce semantic hallucinations, where objects from target classes such as traffic signs and vehicles, as well as man-made light effects, are incorrectly synthesized. These hallucinations significantly degrade downstream performance.
We propose a novel framework that detects and suppresses hallucinations of target-class features during unpaired translation. To detect hallucination, we design a dual-head discriminator that additionally performs semantic segmentation to identify hallucinated content in background regions. To suppress these hallucinations, we introduce class-specific prototypes, constructed by aggregating features of annotated target-domain objects, which act as semantic anchors for each class.
Built upon a Schrödinger Bridge-based translation model, our framework performs iterative refinement, where detected hallucination features are explicitly pushed away from class prototypes in feature space, thus preserving object semantics across the translation trajectory.
Experiments show that our method outperforms existing approaches both qualitatively and quantitatively. On the BDD100K dataset, it improves mAP by {15.5\%} for day-to-night domain adaptation, with a notable {31.7\%} gain for classes such as traffic lights that are prone to hallucinations.
\end{abstract}

\section{Introduction}
\label{sec:intro}
Unpaired image-to-image (I2I) translation has been actively explored in autonomous driving for adapting annotated datasets from daytime to nighttime conditions, supporting downstream tasks such as nighttime object detection~\cite{zhang2023fit, murez2018image, kim2019diversify, huang2024blenda} and semantic segmentation~\cite{jiang2019segmentation, cherian2019sem, peng2023diffusion, wang2024diffusion}.
Its ability to operate without paired images makes it a flexible and scalable approach, particularly given the difficulty of collecting pixel-aligned image pairs.

However, existing unpaired I2I methods, including both GAN-based~\cite{zhu2017unpaired, huang2018multimodal, lee2018diverse, park2020cut} and diffusion-based approaches~\cite{sasaki2021unitddpm, su2023dual, wu2023latent}, typically do not utilize dataset annotations.
As a result, they often fail to preserve objects from target classes such as traffic signs and vehicles. 

To address this, several instance-aware unpaired I2I methods have been proposed that leverage bounding box annotations and class information available during training.
INIT~\cite{shen2019towards} performs separate instance-level and global translations.
DUNIT~\cite{bhattacharjee2020dunit} incorporates object detectors to enhance the modeling of target-class regions.
MGUIT~\cite{jeong2021memory} introduces a memory module to store class-aware styles, while InstaFormer~\cite{kim2022instaformer} leverages a transformer~\cite{vaswani2017attention} encoder to improve instance-level translation.
As shown in Fig.\ref{fig:teaser}, while these methods enhance the translation of annotated foreground objects, they lack mechanisms to constrain the background (unannotated) regions.
Without explicit background control, these models often introduce {hallucinated content} resembling target classes. As illustrated in Table~\ref{tab:bdd}, using such data for object detection can lead to significant degraded performance.

In this paper, we tackle the challenge of semantic hallucinations in day-to-night unpaired translation by introducing a framework that explicitly \textbf{detects} and \textbf{suppresses} hallucinated features related to target classes, i.e., classes annotated in the dataset. Our method builds upon a Schrödinger Bridge-based multi-step translation framework, which progressively refines the image through intermediate stages. This formulation reduces the difficulty of bridging large domain gaps and allows greater diversity in translation.

Target-class hallucinated pixels appear outside of annotated bounding boxes. Pixel-level segmentation masks are required to precisely localize which pixels have target-classes features. However, object detection datasets often provide only bounding box annotations, lacking the segmentation labels. To effectively train a segmentation model, we generate pseudo segmentation masks using a foundation segmentation model~\cite{ravi2024sam}, with the bounding boxes serving as prompts. This enables our model to detect hallucinated pixels beyond annotated object boundaries.
Conventional discriminators often rely on superficial style cues to distinguish real from fake images. In day-to-night translation, this bias encourages generators to insert hallucinated lights and target-like artifacts (e.g., traffic signals, headlights) to mimic nighttime style. To counter this, we integrate target-class hallucination segmentation into a dual-head discriminator, allowing it to detect and penalize semantic inconsistencies rather than relying solely on style cues.

To suppress the detected hallucinations, we construct target-class prototypes by aggregating features of annotated instances in the target domain. These prototypes act as semantic anchors, and hallucinated features detected during intermediate translation steps are pushed away from the anchors via contrastive learning, thus enforcing semantic boundaries between background and foreground.

We evaluate our framework on multiple datasets and tasks.
On the BDD100K~\cite{yu2020bdd100k} dataset, our method improves mAP by \textbf{15.5\%} for day-to-night domain adaptation on object detection, with a \textbf{31.7\%} boost for challenging classes such as traffic lights that are particularly prone to hallucinations.
Our method also achieves state-of-the-art performance on cross-dataset translation and cross-weather tasks.
In summary, our contributions are:
\begin{enumerate}
	\item {\bf Hallucination-Suppression Translation Framework}: detects and suppresses target-class hallucinations in a multi-step unpaired I2I framework, ensuring semantic consistency between translations and annotations.
	\item {\bf Hallucination-Aware Discriminator:} augments the style discriminator with a segmentation head that explicitly predicts target-class hallucinations; pixel-level pseudo masks, obtained by utilizing dataset’s bounding boxes as prompt, supervise this head.
	\item {\bf Prototype-Based Suppression:} regulates hallucinated features by contrasting them against target-class prototypes derived from real target domain features.
	\item {\bf Empirical Validation:}  demonstrates substantially fewer target-class hallucinations and large improvements in downstream detection accuracy across diverse datasets.
\end{enumerate}

\section{Related Work}
\label{sec:related}

\begin{figure*}[t!]
	\centering
	\includegraphics[width=2.1\columnwidth]{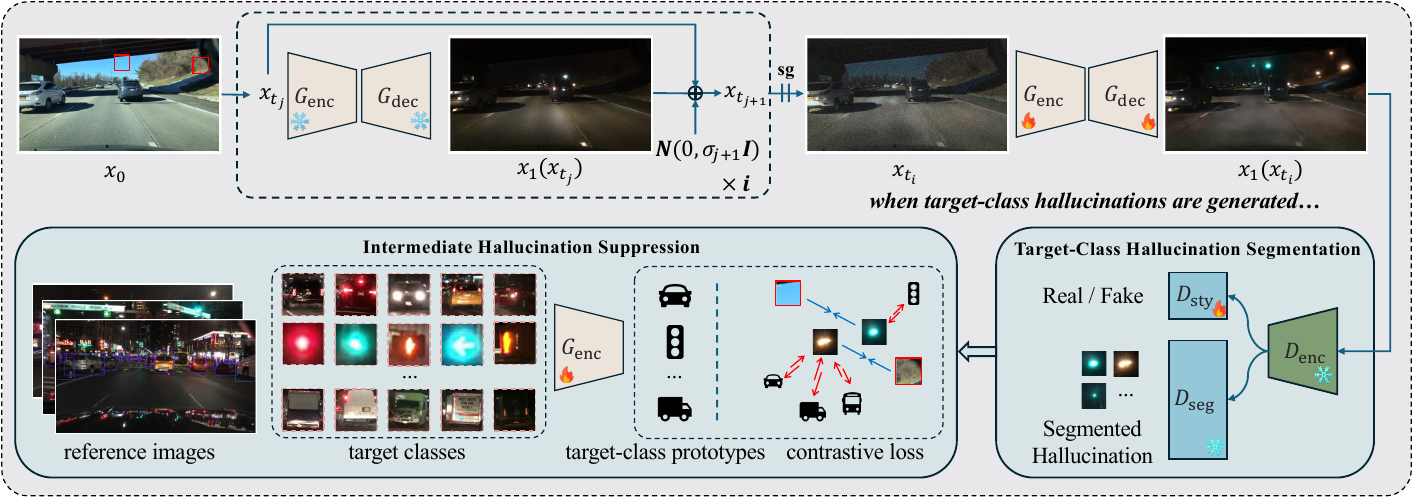} 
	\caption{{Model overview}. 
		Given a source image \( x_0 \), our framework models image-to-image translation as a multi-step transport process that progressively refines the translation.  
		At each step \( j \in [0,i] \), the generator processes the current state \( x_{t_j} \) and predicts an intermediate target image \( x_1(x_{t_j}) \).  
		The next state \( x_{t_{j+1}} \) is obtained by mixing \( x_{t_j} \), \( x_1(x_{t_j}) \), and Gaussian noise.  
		This process continues until \( x_{t_i} \), the final intermediate state before training.  
		Hallucinations in the  \( x_1(x_{t_{i}}) \) are detected by the segmentation head \( D_{\text{seg}} \), while reference images provide annotated objects to generate target-class prototypes of the target domain.  
		Intermediate hallucination suppression then pushes detected hallucinated features away from the prototypes.  
	} 
	\label{fig:overview}
\end{figure*}

\noindent{\bf Image-to-Image Translation } Early I2I methods like Pix2Pix~\cite{isola2017image} achieved strong results with conditional GANs~\cite{goodfellow2014generative}, but required paired data. To relax this constraint, unpaired methods emerged, notably CycleGAN~\cite{zhu2017unpaired}, which introduced cycle-consistency loss, an idea widely adopted~\cite{choi2018stargan, hoffman2018cycada, huang2018multimodal, kim2017disco, lee2018diverse, yi2017dualgan}. However, later work~\cite{wu2024stegogan} revealed that bypassing cycle-consistency can cause semantic errors. 

To reduce complexity, contrastive approaches like CUT~\cite{park2020cut} and F-LSeSim~\cite{wang2022flexit}, inspired by InfoNCE~\cite{oord2018representation}, replaced cycle loss but lack object-level constrain. These models often hallucinate or distort target-class features in large domain shifts (e.g., day-to-night). Diffusion-based models recently advanced unpaired translation: SDEdit~\cite{meng2021sdedit} perturbs and denoises inputs; Cycle-Diffusion~\cite{cyclediffusion} translates via a shared latent space; DDIB~\cite{su2022dual} uses dual diffusion bridges for cycle consistency; InstructPix2Pix~\cite{brooks2023instructpix2pix} and Pix2Pix-Zero~\cite{parmar2023zero} perform text-guided edits. UNSB~\cite{kim2023unpaired} reformulates I2I as multi-step stochastic transport via Schrödinger Bridge, enabling smoother domain shifts.

Despite these advances, most approaches suffer from inversion artifacts~\cite{parmar2024one}, lack realism in structured scenes, and offer no mechanism for hallucination detection or suppression, leading to semantic distortions in challenging domains like day-to-night translation, a difficulty also observed in other nighttime tasks~\cite{lin2024nighthazenighttimeimagedehazing}.

\noindent{\bf Instance-Aware Image-to-Image Translation }
Downstream tasks, such as object detection~\cite{10.1609/aaai.v39i9.33006} and semantic segmentation~\cite{zhang2025mambabridgevisionfoundation},  require reliable translation for domain adaptation.
Several methods have explored instance-aware I2I translation to improve the fidelity of annotated objects by leveraging bounding box supervision for object detection.
InstaGAN~\cite{mo2018instagan} uses segmentation masks to translate only foreground instances while keeping the background unchanged.
INIT~\cite{shen2019towards} translate instances separately with background regions.
DUNIT~\cite{bhattacharjee2020dunit} integrates object detection to preserve the position and scale of instances but fails to prevent hallucinations in unannotated regions. 
Subsequent approaches such as MGUIT~\cite{jeong2021memory} incorporate a memory bank to store class-aware styles, while InstaFormer~\cite{kim2022instaformer} replaces CNN encoders with transformers~\cite{vaswani2017attention} and applies contrastive loss within bounding boxes.

Although these methods improve translation quality inside annotated regions, they overlook semantic inconsistencies outside the boxes.
As a result, under large domain shifts such as day-to-night translation, target-class instances are often hallucinated as background artifacts (e.g., headlights, taillights, traffic lights), degrading the performance of downstream detection models trained on these datasets.

\section{Proposed Method}
Our goal is to learn a mapping from the source domain $ \mathcal{X} $ (daytime) to the target domain $ \mathcal{Y} $ (nighttime) while preserving semantic consistency with annotated target classes.
We assume access to object annotations: bounding boxes and class labels during training, but not at test time.
To address hallucinations related to annotated classes, we propose a Schrödinger Bridge-based translation framework with two tightly integrated components:
(1) Target-Class Hallucination Segmentation, which detects hallucinated pixels within a dual-head discriminator, and
(2) Intermediate Hallucination Suppression, which regulates hallucinated features using class-specific prototypes.

\begin{figure*}[ht]
	\centering
	\includegraphics[width=2.1\columnwidth]{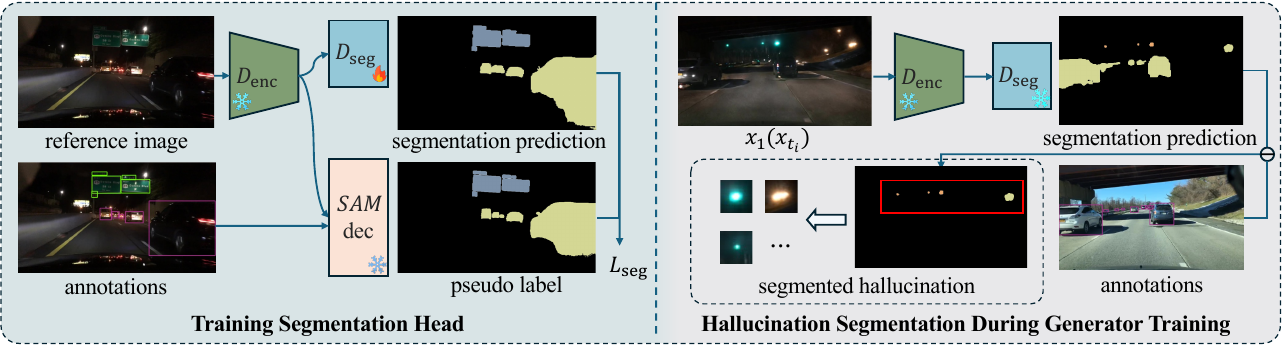}
	\caption{{Discriminator training.} Left panel: The segmentation head \( D_{\text{seg}} \) is trained using SAM2~\cite{ravi2024sam}-generated pseudo labels, with the segmentation loss \( L_{\text{seg}} \). Right panel: In the generator training stage, translated images are evaluated by the discriminator. Hallucinations are segmented by comparing the segmentation prediction with the annotations. The segmented hallucinations will be utilized by the intermediate hallucination suppression. The style head is not shown in this figure.}
	\vspace{-3mm}
	\label{fig:disc}
\end{figure*}

\subsection{Schrödinger Bridge Multi-Step Translation}
Our generator follows the Unpaired Neural Schrödinger Bridge (UNSB~\cite{kim2023unpaired}) framework, which models unpaired image-to-image translation as a sequence of stochastic transport steps between intermediate distributions.  
Instead of directly mapping a source image to its translated counterpart in a single step, the model progressively refines the transformation by constructing a sequence of intermediate states, enabling a smoother and more stable transition from the source domain to the target domain.

As illustrated in Fig.~\ref{fig:overview}, given a source image $x_0 \in \mathcal{X}$, the translation process is formulated as a Markov chain.  
Given a partition $\{t_j\}_{j=0}^N$ of the unit interval $[0,1]$, each step $t_j$ produces an intermediate image $x_{t_j}$.  
The key idea is to first estimate a target-domain image from the current state:
\begin{equation}
	x_1(x_{t_j}) \sim q_\theta(x_1 | x_{t_j}),
\end{equation}
where $q_\theta(x_1 | x_{t_j})$ is a neural network $\theta$ that predicts the target-domain counterpart of $x_{t_j}$.  
This prediction is then used to generate the next state $x_{t_{j+1}}$ as a mixture of the current state, the predicted target image, and Gaussian noise:
\begin{equation}
	x_{t_{j+1}} = s_{j+1} x_1(x_{t_j}) + (1 - s_{j+1}) x_{t_j} + \sigma_{j+1} \epsilon, \quad \epsilon \sim \mathcal{N}(0, I),
\end{equation}
where $s_{j+1} = \frac{t_{j+1} - t_j}{1 - t_j}$ controls the interpolation between $x_{t_j}$ and $x_1(x_{t_j})$, and $\sigma_{j+1}$ determines the level of Gaussian noise.  
This iterative process constructs a trajectory $x_{t_0}, x_{t_1}, \dots, x_{t_N}$, where the final state $x_{t_N} \in \mathcal{Y}$ represents the refined translation in the target domain.

To train the model, in each iteration, the model executes the translation process until a {random} step $t_i$, iteratively generating $x_{t_1}, x_{t_2}, \dots, x_{t_i}$ following the above process.  
The generator then predicts its target-like counterpart $x_1(x_{t_i})$, which is used to compute the training losses and update the parameters $\theta$.
Since $i$ is sampled uniformly from all steps, training losses are not limited to a single frame.
During inference, the model runs the complete Markov chain from $x_0$ to $x_{t_N}$, producing the final translated image.

\subsection{Target-Class Hallucination Segmentation}
\label{sec:disc}

Conventional discriminators assess image realism by focusing on global style, which unintentionally reinforces artifacts that mimic frequent patterns in the target domain.  
In day-to-night translation, for example, the model can hallucinate traffic signals or vehicle headlights, as the discriminator misinterprets these features as essential style elements.

To address this, we design a hallucination-aware discriminator with a dual objective: evaluating global style and detecting hallucinations.  
As illustrated in Fig.~\ref{fig:overview}, the discriminator includes two heads: $D_{\text{sty}}$ for style assessment and $D_{\text{seg}}$ for hallucination segmentation.  
Both heads share a frozen backbone encoder $D_\text{enc}$, based on the hierarchical vision transformer from SAM2~\cite{ravi2024sam, ryali2023hiera}.

\paragraph{Hallucination Segmentation Head} Bounding box annotations can train a detector that enforces coarse semantic consistency during translation, but they can miss the subtle artifacts: a hallucination may appear as a lone car headlight or a hint of a traffic signal. Therefore, we train a segmentation model for pixel-level hallucination detection with only bounding box annotations provided by the datasets.

As shown in Fig.~\ref{fig:disc}, the segmentation head $D_{\text{seg}}$ is implemented as a UNet~\cite{ronneberger2015unet} decoder connected to the encoder $D_\text{enc}$. It is trained on the target domain using SAM2-generated pseudo labels.  
Given a reference image and its annotations $(c_n, x_n, y_n, w_n, h_n)$, where $c_n$ denotes the class and $(x_n, y_n, w_n, h_n)$ the bounding box, SAM2 produces instance-level segmentations.  

We convert these instance masks into a semantic segmentation map $S \in \mathbb{R}^{W \times H}$:
\begin{equation}
	S(w, h) = 
	\begin{cases} 
		c & \text{if } (w, h) \text{ belongs to class } c, \\ 
		0 & \text{if } (w, h) \text{ belongs to background},
	\end{cases}
\end{equation}
where overlapping regions are resolved using confidence scores. This generates a standard semantic pseudo label. 

Although SAM2 is generally robust on nighttime due to its large training data, its masks can still miss fine object boundaries. To improve pseudo-label quality, we enlarge each bounding box by 10\% and using it as a new prompt to obtain a second mask. If the second mask achieves an IoU $>$ 0.9 with the original mask, we keep $S$ as the pseudo label. Otherwise, we discard that region.
Using $S$ as pseudo labels, the segmentation head $D_{\text{seg}}$ is trained to produce pixel-wise predictions on target domain images.

\paragraph{Hallucination Loss In Generator Training} 
When applied to a translated image $x_1(x_{t_i})$, $D_{\text{seg}}$ outputs a semantic segmentation map $S' \in \mathbb{R}^{(C+1) \times W \times H}$, where $C$ is the number of foreground classes and 1 represents the background.  

To penalize hallucinations, we define a hallucination loss $L_{\text{hl}}$ that suppresses the prediction of target classes in unannotated regions:
\begin{equation}
	L_{\text{hl}} = \frac{1}{|S_{bg}|} \sum_{(w,h) \in S_{bg}} \sum_{c=1}^{C} \left( \text{softmax}(\hat{S})_{cwh} \right)^2,
\end{equation}
where $\text{softmax}(\hat{S})_{cwh}$ is the probability of class $c$ at pixel $(w, h)$, and $S_{bg}$ is the set of background pixels.
This loss reduces false activations of target classes in unannotated regions, which is critical in structured environments like driving scenes.
Without this constraint, background areas may be incorrectly translated into objects such as headlights or traffic lights, degrading the utility of the translated images for downstream tasks.

\begin{figure*}[ht!]
	\centering
	\begin{overpic}[width=2.1\columnwidth]{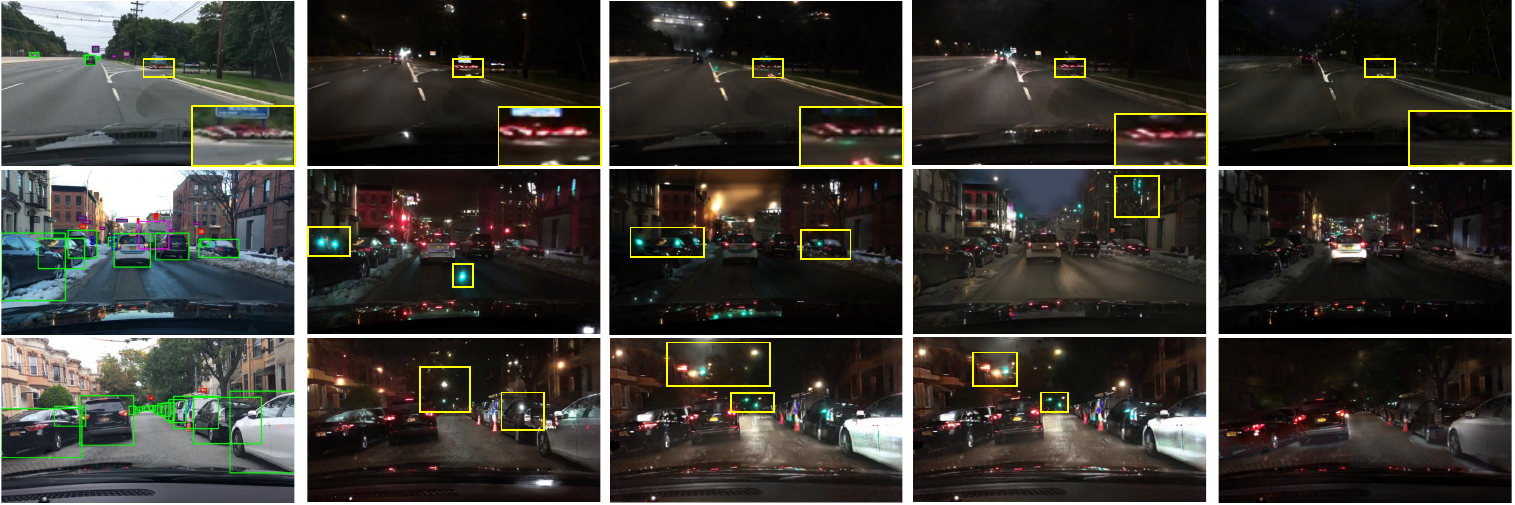} % Replace with your image path
		\put(7, -1.4){\small (a) Input}
		\put(24, -1.4){\small (b) InstaFormer}
		\put(46, -1.4){\small (c) MGUIT}
		\put(67, -1.4){\small (d) UNSB}
		\put(87, -1.4){\small (e) Ours}
	\end{overpic}
	\vspace{0.8em}
	\caption{{Qualitative comparison} of day-to-night translation results on the BDD100K dataset.
		Yellow boxes highlight regions containing target-class hallucinations or inconsistencies with the original annotations, which may introduce label noise. Our method (e) preserves both the realistic nighttime style and semantic consistency with the input annotations, whereas existing state-of-the-art methods (b–d) often produce hallucinated objects or fail to align with the source semantics.
	}
	\vspace{-4mm}
	\label{fig:compare}
\end{figure*}
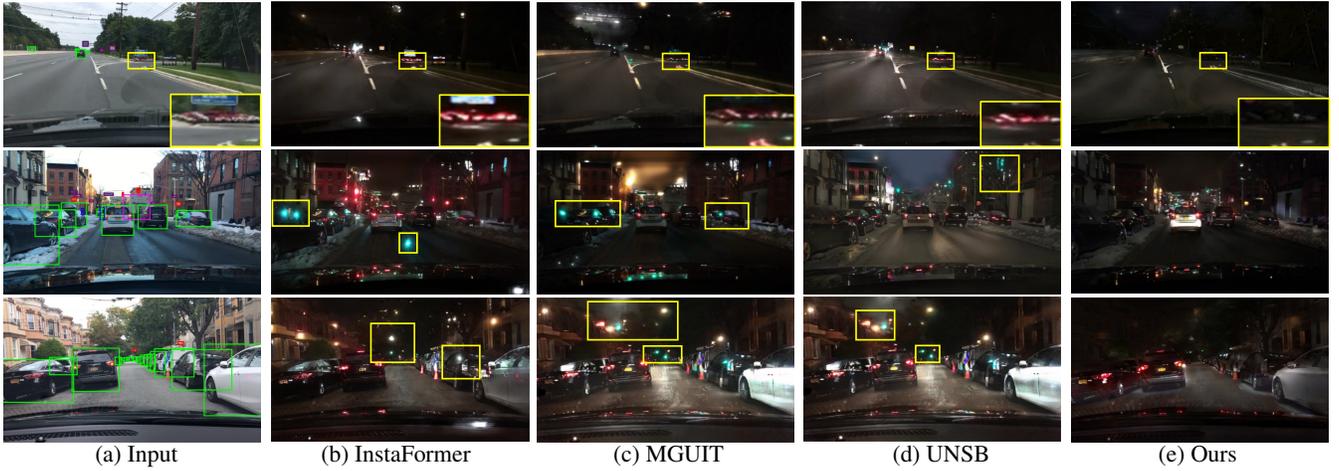

\subsection{Intermediate Hallucination Suppression}
\label{sec:halluc}

Existing instance-aware methods~\cite{mo2018instagan,shen2019towards,bhattacharjee2020dunit,jeong2021memory,kim2022instaformer} focus on refining features within bounding boxes using category information to guide object-level translation.  
However, regions outside these bounding boxes are equally important. They represent the background and should exhibit features that are clearly distinct from those of foreground objects.  
Ignoring this distinction causes background regions to drift toward target-class hallucinations.

To address this, we treat foreground features from reference images as negative examples for background regions. This encourages the generator to maintain a clear separation between target classes and background during translation.
Directly using instance features from training batches presents two challenges.  
First, some classes may be absent from individual batches, introducing instability if batch-specific features are used.  
Second, high intra-class variability makes it difficult to enforce a consistent separation between foreground and background.
To overcome these issues, we maintain class-wise prototypes. They are mean feature representations aggregated across batches, which serve as stable anchors for each class.

\paragraph{Prototype Formulation }  
Let $\mathcal{C} = \{c_1, c_2, \dots, c_k\}$ be the set of foreground classes, with each class $c_i$ associated with a prototype $p_{c_i}$.  
For each instance of $c_i$ in the reference image, we extract a set of feature vectors $\{f_{c_i}^{(j)}\}$ from the encoder $G_{\text{enc}}$ within the bounding box.  
The prototype $p_{c_i}$ is computed by averaging these features:
\begin{equation}
	p_{c_i} = \frac{1}{N_{c_i}} \sum_{j=1}^{N_{c_i}} f_{c_i}^{(j)},
\end{equation}
where $N_{c_i}$ is the total number of feature vectors extracted for class $c_i$ in the batch.

To keep prototypes up to date, we apply an exponential moving average (EMA) update across training batches:
\begin{equation}
	p_{c_i}^{\text{new}} = \alpha \cdot p_{c_i} + (1 - \alpha) \cdot \frac{1}{N_{\text{batch}}} \sum_{j=1}^{N_{\text{batch}}} f_{c_i}^{(j)},
\end{equation}
where $\alpha$ is a momentum parameter.

\noindent\textbf{Hallucinated Feature Suppression }  
To identify hallucinated pixels $S_{\text{hl}}$, we detect background pixels that resemble foreground classes. A pixel $(w, h)$ is considered hallucinated if the highest softmax score across foreground classes exceeds that of the background:
\begin{equation}
	\max_{c \in \{1, \dots, C\}} \text{softmax}(\hat{S})_{cwh} > \text{softmax}(\hat{S})_{0,wh},
\end{equation}
where $\text{softmax}(\hat{S})_{cwh}$ is the predicted probability for class $c$, and $\text{softmax}(\hat{S})_{0,wh}$ is for background.

To suppress these hallucinations, we apply an InfoNCE loss~\cite{oord2018representation} using contrastive learning.  
The anchor $\hat{\mathbf{v}} = G_{\text{enc}}(\hat{y})$ is the feature at the hallucinated pixel in the translated image $\hat{y}$. The positive sample $\mathbf{v}^+ = G_{\text{enc}}(x)$ is the feature at the corresponding pixel in the source image $x$, and the negatives $\mathbf{v}^-$ are features from other locations. Class prototypes are used as additional negatives to enhance separation. The hallucination suppression loss is defined as:
\begin{multline}
	\mathcal{L}_{\text{supp}}(\hat{\mathbf{v}}, \mathbf{v}^+, \{\mathbf{v}^-_n\}_{n=1}^N) = \\
	-\log \frac{\exp(\hat{\mathbf{v}} \cdot \mathbf{v}^+ / \tau)}{\exp(\hat{\mathbf{v}} \cdot \mathbf{v}^+ / \tau) + \sum_{n=1}^{N} \exp(\hat{\mathbf{v}} \cdot \mathbf{v}^-_n / \tau) + \text{PDist}}
\end{multline}
where $\text{PDist} = \sum_{c=1}^{C} \exp(\hat{\mathbf{v}} \cdot {p}_{c_i} / \tau)$. 
Variable $\tau$ is the temperature, $N$ is the number of negative samples, and $C$ is the number of prototypes.  
$\text{PDist}$ aggregates the distances between the hallucinated feature and prototypes, enforcing feature separation between background and target classes.

\begin{table*}[ht!]
	\centering
	
	\resizebox{\textwidth}{!}{%
		\small
		\begin{tabular}{l|c|ccccccccc}
			\toprule
			\textbf{Methods} & \textbf{mAP} & \textbf{Person} & \textbf{Car} & \textbf{Rider} & \textbf{Bus} & \textbf{Truck} & \textbf{Bike} & \textbf{Motor} & \textbf{T. Light} & \textbf{T. Sign} \\
			\midrule
			{Lower Bound}  & 13.75 & 12.99 & 25.21 & 8.71  & 18.93 & 15.73 & 8.45  & 6.88  & 8.28 & 18.55 \\
			{Upper Bound}  & 17.86 & 14.43 & 32.59 & 10.25 & 26.50 & 23.83 & 8.96  & 8.43  & 11.93 & 23.83 \\
			\hline
			DRIT & 12.68 & 11.84 & 25.16 & 8.24 & 14.89 & 15.46 & 9.05 & 3.84 & 6.49 & 19.18 \\
			CycleGAN & 13.16 & 12.52 & 26.08 & 7.46 & 17.36 & 16.01 & 9.46 & 5.70 & 4.60 & 19.29 \\
			CUT &  14.10 & 14.13 & 28.31 & 8.27 & 20.22 & 16.21 & 9.96 & 5.29 & 5.36 & 19.19 \\
			\hline
			%	SDEdit & 11.32 & 12.21 & 23.44 & 7.35 & 17.29 & 14.86 & 9.27 & 5.38 & 3.92 & 13.18 \\
			Cycle-Diffusion & 13.62 & 13.54 & 26.49 & 8.31 & 20.32 & 17.02 & 10.11 & 6.22 & 4.91 & 17.88 \\
			DDIB & 10.67 
			& 10.75 
			& 20.40 
			& 6.66 
			& 15.83 
			& 13.42 
			& 7.87 
			& 4.82 
			& 3.80 
			& 12.49 \\
			InstructPix2Pix & 10.84 & 11.89 & 21.76 & 6.82 & 15.33 & 13.71 & 8.56 & 4.97 & 3.45 & 12.07 \\
			%	Pix2Pix-Zero & 11.45 & 12.14 & 22.34 & 7.12 & 16.42 & 14.26 & 8.98 & 5.12 & 3.78 & 12.63 \\
			UNSB & 14.27 & 14.65 & 28.35 & 8.95 & 22.83 & 17.14 & 9.68 & 6.02 & 5.93 & 14.88 \\
			\hline
			DUNIT & 14.87 & 14.49 & 28.03 & 9.05 & 23.38 & 19.47 & 10.15 & 6.38 & 4.43 & 18.42 \\
			MGUIT & 15.08 & 14.52 & 27.48 & 8.79 & 23.41 & 19.08 & 10.93 & 6.53 & 6.18 & 18.83 \\
			InstaFormer & 14.93 & 14.04 & 27.25 & 8.58 & 21.67 & 19.48 & 10.98 & 7.83 & 6.33 & 18.19 \\
			\hline
			\rowcolor{gray!15}
			\textbf{Ours}  & \textbf{17.40} & \textbf{15.35} & \textbf{30.01} & \textbf{10.92} & \textbf{26.02} & \textbf{23.48} & \textbf{11.74} & \textbf{8.44} & \textbf{8.55} & \textbf{22.01} \\
			\bottomrule
		\end{tabular}%
		
	}
	\caption{{Comparison on the BDD100K Day-to-Night Translation Task.} The Lower Bound represents the detector trained in the daytime, while the Upper Bound represents the detector trained in the nighttime. All methods are trained on daytime images and translated nighttime images and tested at nighttime. Reported AP is the averaged AP over IoUs 0.5 to 0.95.}
	\label{tab:bdd}
\end{table*}

\subsection{Loss Functions}
In addition to $\mathcal{L}_{\text{hl}}$ and $\mathcal{L}_{\text{supp}}$, we optimize our model using a weighted combination of several loss terms.

\noindent\textbf{Adversarial Loss }  To encourage the translated images consistent with the target style, we adopt an adversarial loss $\mathcal{L}_{\text{adv}}$ through with the style head $D_\text{sty}$ in the discriminator.

\noindent\textbf{Schrödinger Bridge Loss }  
To encourage semantic alignment and promote diversity in translations, we adopt the Schrödinger Bridge loss from UNSB~\cite{kim2023unpaired}. At each step $t_i$, this loss minimizes the transport cost between intermediate and target samples, while maximizing entropy to allow diverse and plausible outputs:
\begin{equation}
	\begin{split}
		\mathcal{L}_{\text{SB}} &= \mathbb{E}_{q_\theta(x_{t_i}, x_1)}\left[\|x_{t_i} - x_1\|^2\right] \\
		&\quad - 2\tau(1 - t_i) \cdot H(q_\theta(x_{t_i}, x_1)),
	\end{split}
\end{equation}
where $\tau$ controls the noise level and $H(\cdot)$ denotes entropy.

\noindent\textbf{Regularization Loss }  
To ensure consistency between the input image and the translated output, we apply feature-level regularization on the generator using contrastive learning~\cite{oord2018representation}. This loss is implemented using a two-layer MLP on top of encoder features, promoting semantic alignment while suppressing hallucinations:
\begin{itemize}
	\item $\mathcal{L}_{\text{cont}}$ operates over all features, aligning corresponding spatial locations, similar to CUT~\cite{park2020cut}.
	\item $\mathcal{L}_{\text{supp}}$ is restricted to hallucinated regions and incorporates class prototypes as negative samples.
\end{itemize}

\noindent\textbf{Segmentation Loss }  
To supervise the segmentation head $D_{\text{seg}}$ using SAM2-generated pseudo labels, we apply a softmax cross-entropy loss.

\noindent\textbf{Total Loss}  
The final objective is a weighted sum of the individual components:
\[
\mathcal{L}_{\text{total}} = \lambda_1 \mathcal{L}_{\text{adv}} +  \lambda_2 \mathcal{L}_{\text{SB}}+  \lambda_3 \mathcal{L}_{\text{seg}} + \lambda_4 \mathcal{L}_{\text{cont}} + \lambda_5 \mathcal{L}_{\text{supp}} + \lambda_6 \mathcal{L}_{\text{hl}},
\]
where each $\lambda_i$ is the scaler of its respective loss term.

\section{Experiments}

\noindent\textbf{Implementation Details }  
All experiments are conducted on eight RTX 3090 GPUs.
Our framework is trained for 100 epochs using the Adam optimizer~\cite{kingma2014adam}, with a batch size of 8 and an initial learning rate of 0.0001. A step decay scheduler adjusts the learning rate.
Loss weights are set as $\lambda_1$ to $\lambda_5 = 1$, and $\lambda_6 = 0.2$.
We use the Hiera vision transformer (Hiera-T)~\cite{ryali2023hiera} as the frozen shared encoder.
For object detection, we adopt Faster R-CNN~\cite{ren2016faster} with a ResNet-50 backbone~\cite{he2016deep} and FPN head, implemented in Detectron2~\cite{wu2019detectron2}.
The detector is pretrained on ImageNet~\cite{deng2009imagenet} and then trained for 40,000 iterations.

\begin{table*}[t!]
	\centering
	\resizebox{\textwidth}{!}{%
		\small
		\begin{tabular}{l|c|ccccccccc}
			\toprule
			\textbf{Methods} & \textbf{mAP} & \textbf{Person} & \textbf{Car} & \textbf{Rider} & \textbf{Bus} & \textbf{Truck} & \textbf{Bike} & \textbf{Motor} & \textbf{T. Light} & \textbf{T. Sign} \\
			\hline
			{Lower Bound}  & 13.75 & 12.99 & 25.21 & 8.71  & 18.93 & 15.73 & 8.45  & 6.88  & 8.28 & 18.55 \\
			\hline
			{w/o \( \mathcal{L}_{\text{hl}} \) \& \( \mathcal{L}_{\text{supp}} \)} 
			& 14.11 & 14.08 & 27.14 & 8.31 & 19.66 & 18.42 & 9.66 & 4.48 & 5.48 & 19.82 \\
			{w/o \( \mathcal{L}_{\text{supp}} \)} 
			& 15.55 & 14.80 & 27.69 & 9.00 & 22.15 & 20.10 & 11.84 & 8.42 & 7.01 & 18.94 \\
			{w/o \( \mathcal{L}_{\text{hl}} \)} 
			& 16.43 & 15.26 & 29.65 & 10.22 & 25.49 & 20.42 & 11.12 & 7.97 & 7.45 & 20.34 \\
			\hline
			\rowcolor{gray!15} 
			\textbf{All Components} 
			& \textbf{17.40} & \textbf{15.35} & \textbf{30.01} & \textbf{10.92} & \textbf{26.02} & \textbf{23.48} & \textbf{11.74} & \textbf{8.44} & \textbf{8.55} & \textbf{22.01} \\
			\bottomrule
		\end{tabular}%
	}
	\caption{{Ablation study on the mAP for BDD100K day-to-night domain adaptation.} \( \mathcal{L}_{\text{hl}} \) refers to the hallucination segmentation loss, and \( \mathcal{L}_{\text{supp}} \) denotes the hallucination feature suppression loss.}
	\label{tab:ablation}
\end{table*}

\paragraph{Datasets }   
We evaluate our method across two benchmarks:
(1) BDD100K~\cite{yu2020bdd100k}:  
we evaluate day-to-night domain adaptation using the BDD100K dataset, which presents strong appearance shifts and frequent hallucinations in nighttime scenes.
The training set includes 36,728 day and 27,971 night images; 3,929 night images with bounding box annotations are used for evaluation.
(2) {KITTI}$\rightarrow${Cityscapes}:  
we translate KITTI (7.5k train / 7.5k test) to the Cityscapes target domain (5k fine annotations) and evaluate detection on the four classes: \textit{person}, \textit{car}, \textit{truck}, and \textit{bicycle}.
We follow prior work~\cite{kim2022instaformer}, using 85\% for training and 15\% for testing, and conduct weather translation experiments.

\begin{table}[t]
	\centering
	\renewcommand{\arraystretch}{0.6}
	
	\sisetup{
		detect-weight, 
		mode=text      
	}

	\begin{tabular}{l S[table-format=2.1] S[table-format=2.1] S[table-format=2.1] S[table-format=2.1] S[table-format=2.1]}
		\toprule
		\textbf{Methods} & {\textbf{mAP}} & {\textbf{Person}} & {\textbf{Car}} & {\textbf{Truck}} & {\textbf{Bike}} \\
		\midrule
		DT          & 31.2 & 28.5 & 40.7 & 25.9 & 29.7 \\
		DAF         & 38.5 & 39.2 & 40.2 & 25.7 & 48.9 \\
		DARL        & 45.3 & 46.4 & 58.7 & 27.0 & 49.1 \\
		DAOD        & 46.1 & 47.3 & 59.1 & 28.3 & 49.6 \\
		DUNIT       & 54.1 & 60.7 & 65.1 & 32.7 & 57.7 \\
		MGUIT       & 54.6 & 58.3 & 68.2 & 33.4 & 58.4 \\
		InstaFormer & 55.5 & 61.8 & \bfseries 69.5 & 35.3 & 55.3 \\
		\midrule 
		\rowcolor{gray!15}
		\textbf{Ours} & \bfseries 57.4 & \bfseries 62.6 & \bfseries 69.5 & \bfseries 37.8 & \bfseries 59.8 \\
		\bottomrule
	\end{tabular}
	\caption{{KITTI → Cityscapes Domain Adaptation.} Per-class comparison is shown.}
	\label{tab:comparison_map} % <-- Moved label for correct referencing
\end{table}

\subsection{Domain Adaptive Object Detection}

To evaluate the effectiveness and reliability of translated data, we conduct domain adaptation experiments on object detection across two settings: day-to-night adaptation and cross-dataset adaptation.  
In these translation tasks, hallucinations can degrade detection performance by introducing label noise and false positives, particularly for visually ambiguous or light-sensitive categories.

\paragraph{Day-to-Night Experiment }  
We evaluate our method on the BDD100K dataset~\cite{yu2020bdd100k}, a challenging benchmark due to the large visual gap between day and night scenes. Existing I2I methods often introduce background hallucinations (e.g., light artifacts), which lead to false detections and reduced semantic consistency during training.
We define the \textbf{Lower Bound} as the performance of a detector trained on real daytime images (36,728), and the \textbf{Upper Bound} as that of a model trained on real nighttime images (27,971).
We compare against unpaired I2I models~\cite{zhu2017unpaired,park2020cut,parmar2024one}, including Diffusion-based methods~\cite{meng2021sdedit,cyclediffusion,su2022dual,brooks2023instructpix2pix,parmar2023zero,kim2023unpaired} and instance-aware models~\cite{bhattacharjee2020dunit,jeong2021memory,kim2022instaformer}, training all detectors on a mix of real daytime and translated nighttime images.
Input images are resized to $512 \times 512$ in this experiment.

As shown in Table~\ref{tab:bdd}, our method improves mAP by 13.1\% over the previous state-of-the-art and approaches the Upper Bound.
It even exceeds the Upper Bound in several categories: truck, bike, rider, and person, an outcome not seen in prior work.
Some baselines fall below the Lower Bound, indicating that their translations harm detection.
In contrast, our method yields consistent gains across classes. For instance, while all other models degrade on traffic light, ours achieves a 30.7\% boost over the previous best.

\paragraph{KITTI to Cityscapes Experiment }  
Following prior work~\cite{bhattacharjee2020dunit,jeong2021memory,kim2022instaformer}, we evaluate our method on the cross-dataset adaptation task from KITTI~\cite{geiger2012we} to Cityscapes~\cite{cordts2016cityscapes}, and input images are resized to $416 \times 416$ in this experiment.
We adopt the evaluation setup from~\cite{kim2022instaformer} and compare against domain adaptation methods~\cite{inoue2018cross,saito2019strong,kim2019diversify,lopez2019domain} and instance-aware I2I approaches~\cite{bhattacharjee2020dunit,jeong2021memory,kim2022instaformer}.
Our method consistently achieves the highest accuracy across most object classes, demonstrating that our translated datasets offer stronger support for cross-domain object detection.

\subsection{Image Quality Evaluation}

\paragraph{Qualitative Evaluation }  
We compare visual results on the day-to-night translation task using the BDD100K dataset. As shown in Fig.~\ref{fig:compare}, our method produces visually superior outputs, especially in key elements such as traffic lights, taillights, and headlights. These features are crucial for alignment with annotations, as they serve as key indicators for target class detection. Our model generates more realistic lighting, avoiding the overexposed or dim effects often seen in baselines and better preserves the semantic boundaries of foreground objects. It also effectively suppresses background hallucinations, maintaining semantic fidelity. By accurately transferring style while preserving object structure, our method produces translations better suited for downstream tasks that depend on object integrity.

\subsection{Ablation Study}

To evaluate the contribution of each component in our framework, we conduct ablation experiments on the BDD100K day-to-night domain adaptation task.
The goal is to isolate the effects of the hallucination segmentation loss $ \mathcal{L}_{\text{hl}} $ and the feature suppression loss $ \mathcal{L}_{\text{supp}} $ on the translation quality and their impact on the downstream task.

\noindent\textbf{Effect of Hallucination Suppression Components }  
As shown in Table~\ref{tab:ablation}, removing both $ \mathcal{L}_{\text{hl}} $ and $ \mathcal{L}_{\text{supp}} $ yields only a slight improvement over the lower bound (14.11 vs. 13.75 mAP), indicating that hallucination control is critical.
The best performance, 17.40 mAP, is achieved when both components are enabled, delivering the highest accuracy across nearly all classes, with substantial gains in challenging categories like traffic lights and traffic signs.
Notably, average precision for traffic lights only exceeds the lower bound when both losses are applied, underscoring the importance of jointly suppressing and detecting hallucinated features.

\begin{figure}[t]
	\centering
	\includegraphics[width=1\columnwidth]{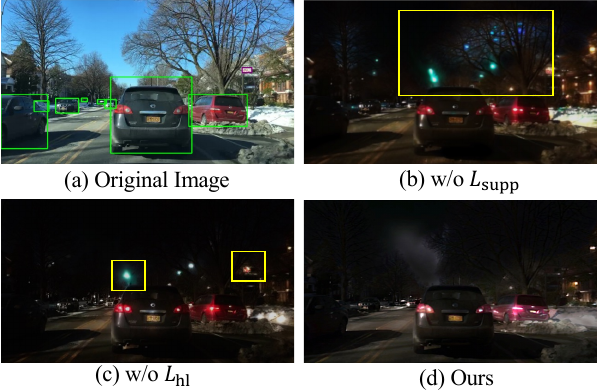}
	\caption{ \textbf{Ablation study}. The fewest hallucinations are observed when both components are incorporated.}
	\label{fig:ablation}
\end{figure}

\noindent\textbf{Qualitative Impact }  
Visual comparisons in Fig.~\ref{fig:ablation} further support these findings. The full model produces translations with fewer hallucinations and greater consistency in object placement, lighting, and style.
These results show that $ \mathcal{L}_{\text{hl}} $ and $ \mathcal{L}_{\text{supp}} $ work together to improve translation fidelity and boost downstream detection performance.

\section{Conclusion}
Target-class hallucinations are a critical yet overlooked issue in unpaired image-to-image translation, as they significantly impair downstream tasks. To address this, we propose a framework that explicitly detects and suppresses hallucinations in annotated classes. Our method combines a hallucination-aware discriminator with a prototype-based suppression mechanism to separate foreground objects from background regions during translation. By utilizing target-class prototypes, it preserves semantic consistency and prevents background artifacts from imitating annotated classes. Extensive experiments across multiple datasets show consistent gains in translation quality and downstream object detection performance.

%%%%%%%%%%%%%%%%%%%%%%%%%%%%%%%%
\bibliography{main}

\end{document}